\pdfoutput=1

\documentclass[11pt]{article}

\usepackage{acl}

\usepackage[T1]{fontenc}

\usepackage[utf8]{inputenc}

\usepackage{hyperref}

\usepackage{inconsolata}
\usepackage{latexsym}
\usepackage{bbm}
\usepackage{microtype}
\usepackage{times,latexsym}
\usepackage{url}
\usepackage{microtype}
\usepackage{graphicx}
\usepackage{multirow}
\usepackage[fixed]{fontawesome5}

\usepackage{multicol}
\usepackage{tabularx}
\usepackage{wrapfig}
\usepackage{booktabs}
\usepackage{subcaption}
\usepackage{kantlipsum}
\usepackage{hhline}
\usepackage{hyperref}
\usepackage{longtable}
\usepackage{tikz}
\usepackage{url}
\usepackage[toc,page,header]{appendix}
\usepackage{titletoc}

\usepackage{xcolor, soul}
\PassOptionsToPackage{dvipsnames}{xcolor}
\definecolor{HallRedShade}{HTML}{fdf1ec}
\definecolor{HallRedText}{HTML}{b76039}
\sethlcolor{HallRedShade}

\usepackage{etoolbox}
\usepackage[inline]{enumitem}
\usepackage{rotating}
\usepackage{amsmath,amsthm,mathtools,amsfonts,amssymb}

 \usepackage{colortbl} 
\usepackage{longtable}
\usepackage{tabu}
\usepackage{longtable}
\usepackage{arydshln}
\makeatletter
\def\adl@drawiv#1#2#3{%
        \hskip.5\tabcolsep
        \xleaders#3{#2.5\@tempdimb #1{1}#2.5\@tempdimb}%
                #2\z@ plus1fil minus1fil\relax
        \hskip.5\tabcolsep}
\newcommand{\cdashlinelr}[1]{%
  \noalign{\vskip 1.3pt
           \global\let\@dashdrawstore\adl@draw
           \global\let\adl@draw\adl@drawiv}
  \cdashline{#1}[.4pt/2pt]
  \noalign{\global\let\adl@draw\@dashdrawstore
           \vskip 1.3pt}}
\makeatother
\usepackage[most]{tcolorbox}

\newcommand{\dashifted}{\raisebox{0.5\depth}{\tiny$\downarrow$}}

\newcommand{\uashifted}{\raisebox{0.5\depth}{\tiny$\uparrow$}}

\definecolor{c1}{cmyk}{0,0.6175,0.8848,0.1490} 
\definecolor{c2}{cmyk}{0.1127,0.6690,0,0.4431}
\definecolor{c3}{cmyk}{0.3081,0,0.7209,0.3255} 
\definecolor{c4}{cmyk}{0.6765,0.2017,0,0.0667}
\definecolor{c5}{cmyk}{0,0.8765,0.7099,0.3647} 
\definecolor{forestgreen}{HTML}{397727}

\definecolor{lightblue}{RGB}{212, 235, 255}
\definecolor{lightorange}{RGB}{255, 204, 168}
\definecolor{lightyellow}{RGB}{255, 255, 168}
\definecolor{lightgreen}{RGB}{224, 242, 213}
\definecolor{lightred}{RGB}{249,202,202}
\definecolor{lightgray}{RGB}{230,230,230}

\newtcbox{\hlprimarytab}{on line, rounded corners, box align=base, colback=lightgreen,colframe=white,size=fbox,arc=3pt, before upper=\strut, top=-2pt, bottom=-4pt, left=-2pt, right=-2pt, boxrule=0pt}
\newtcbox{\hlsecondarytab}{on line, box align=base, colback=lightred,colframe=white,size=fbox,arc=3pt, before upper=\strut, top=-2pt, bottom=-4pt, left=-2pt, right=-2pt, boxrule=0pt}

\newcommand{\da}[1]{{\scriptsize\hlprimarytab{\dashifted{#1}}}}
\newcommand{\ua}[1]{{\scriptsize\hlsecondarytab{\uashifted{#1}}}}
\newcommand{\uag}[1]{{\scriptsize\hlprimarytab{\uashifted{#1}}}}

\newcommand{\dab}[1]{{\scriptsize\hlsecondarytab{\dashifted{#1}}}}

\title{Enhanced Hallucination Detection in Neural Machine Translation through Simple Detector Aggregation}

\author{ \bf Anas Himmi$^{1}$ \, Guillaume Staerman$^{2}$ \, Marine Picot$^{3}$\\ \bf Pierre Colombo$^{1,4}$ \, Nuno M. Guerreiro$^{1,5,6,7}$\\ 
$^{1}$MICS, CentraleSupélec, Universite Paris-Saclay, Paris, France, \\
$^{2}$Universite Paris-Saclay, Inria, CEA, Palaiseau, France, \\
$^{3}$digeiz, Paris, France,
$^{4}$Equall, Paris, France,
$^{5}$Instituto de Telecomunicações, Lisbon, Portugal \\
$^{6}$Unbabel, Lisbon, Portugal,
$^{7}$Instuto Superior Técnico, University of Lisbon, Portugal
}

\begin{document}
\maketitle 
\begin{abstract}
Hallucinated translations pose significant threats and safety concerns when it comes to practical deployment of machine translation systems. Previous research works have identified that detectors exhibit complementary performance --- different detectors excel at detecting different types of hallucinations. In this paper, we propose to address the limitations of individual detectors by combining them and introducing a straightforward method for aggregating multiple detectors. Our results demonstrate the efficacy of our aggregated detector, providing a promising step towards evermore reliable machine translation systems.
\end{abstract}

\section{Introduction}
Neural Machine Translation (NMT) has become the dominant methodology for real-world machine translation applications and production systems. As these systems are deployed \textit{in-the-wild} for real-world usage, it is ever more important to ensure that they are highly reliable. While NMT systems are known to suffer from various pathologies~\citep{koehn-knowles-2017-six}, the most severe among them is the generation of translations that are detached from the source content, typically known as \textit{hallucinations}~\citep{raunak-etal-2021-curious, guerreiro2022looking}. Although rare, particularly in high-resource settings, these translations can have dramatic impact on user trust \cite{perez2022red}. As such, researchers have worked on (i)~methods to reduce hallucinations either during training-time or even inference time~\citep{xiao-wang-2021-hallucination, guerreiro2022looking, dale2022detecting, sennrich2024mitigating}, and alternatively, (ii)~the development of highly effective on-the-fly hallucination detectors~\citep{guerreiro2022looking, guerreiro2022optimal, dale2022detecting} to flag these translations before they reach end-users. In this paper, we will focus on the latter.

One immediate way to approach the problem of hallucination detection is to explore high-quality \textit{external} models that can serve as proxies to measure detachment from the source content, e.g., quality estimation (QE) models such as \textsf{CometKiwi}~\citep{rei-etal-2022-cometkiwi}, or cross-lingual sentence similarity models like \textsf{LASER}~\citep{artetxe-schwenk-2019-massively} and \textsf{LaBSE}~\citep{feng-etal-2022-language}. Intuitively, extremely low-quality translations or translations that are very dissimilar from the source are more likely to be hallucinations. And, indeed, these detectors can perform very effectively as hallucination detectors~\citep{guerreiro2022looking, dale2022detecting}. Alternatively, another effective approach is to leverage \textit{internal} model features such as attention maps and sequence log-probability~\citep{guerreiro2022looking, guerreiro2022optimal, dale2022detecting}. The assumption here is that when translation models generate hallucinations, they may reveal anomalous internal patterns that can be highly predictive and useful for detection, e.g., lack of contribution from the source sentence tokens to the generation of the translation~\citep{ferrando-etal-2022-towards}. Most importantly, different detectors exhibit complementary properties. 
For instance, oscillatory hallucinations --- translations with anomalous repetitions of phrases or $n$-grams~\citep{raunak-etal-2021-curious} --- are readily identified by \textsf{CometKiwi}, while detectors based on low source contribution or sentence dissimilarity struggle in this regard. Therefore, there is an inherent trade-off stemming from the diverse anomalies different detectors excel at.

In this paper, we address this trade-off by proposing a simple yet highly effective method to aggregate different detectors to leverage their complementary strengths. Through experimentation in the two most widely used hallucination detection benchmarks, we show that our method consistently improves detection performance.

\noindent Our key contributions are can be summarized as follows:
\begin{itemize}
    \item  We propose  \textsf{STARE}, an unsupervised \underline{S}imple de\underline{T}ectors \underline{A}gg\underline{R}\underline{E}gation method that achieves state-of-the-art performance well on two hallucination detection benchmarks.
\item We demonstrate that our consolidated detector can outperform single-based detectors with as much as aggregating two complementary detectors. Interestingly, our results suggest that internal detectors, which typically lag behind external detectors, can be combined in such a way that they outperform the latter.
\end{itemize}
We release our code and scores to support future research and ensure reproducibility.\footnote{Code is available here: \url{https://github.com/AnasHimmi/Hallucination-Detection-Score-Aggregation}.}

\section{Detectors Aggregation Method}

\subsection{Problem Statement}
\paragraph{Preliminaries.} Consider a vocabulary $\Omega$ and let $(X, Y)$ be a random variable taking values in $\mathcal{X} \times \mathcal{Y}$, where $\mathcal{X} \subseteq \Omega$ represents translations and $\mathcal{Y} = \{0, 1\}$ denotes labels indicating whether a translation is a hallucination ($Y=1$) or not ($Y=0$). The joint probability distribution of $(X, Y)$ is $P_{XY}$.

\paragraph{Hallucination detection.} The goal of hallucination detection is to classify a given translation $x \in X$ as either an expected translation from the distribution $P_{X|Y=0}$ or as a hallucination from $P_{X|Y=1}$. This classification is achieved by a binary decision function $g: X \rightarrow {0, 1}$, which applies a threshold $\gamma \in \mathbb{R}$ to a hallucination score function $s: X \rightarrow \mathbb{R}$. The decision function is defined as:
\begin{equation*}
    g(x) = \left\{\begin{array}{ll}
1 & \text {if } s(x) > \gamma, \\
0 & \text{otherwise}.
\end{array}\right.
\label{eq:hallucination_detection}
\end{equation*}

The objective is to create an hallucination score function $s$ that effectively distinguishes hallucinated translations from other translations.


\paragraph{Aggregation.} Assume that we have several hallucination score detectors\footnote{We use the notation $\{s_k\}_{k=1}^{K}$ to represent a set consisting of $K$ hallucination detectors, where each $s_k$ is a function mapping from $\mathcal{X}$ to $\mathbb{R}$.}. When evaluating a specific translation $x'$, our goal is to combine the scores from the single detectors into a single, more reliable score that outperforms any of the individual detectors alone. Formally, this aggregation method, denoted as $\operatorname{Agg}$, is defined as follows:
\begin{align*}
     \operatorname{Agg} : \quad \quad \quad \quad \quad \mathbb{R}^{K}  & \rightarrow \mathbb{R} \nonumber\\
      \{ s_k(x')\}_{k=1}^{K} &  \rightarrow \operatorname{Agg}\bigg (\{ s_k\}_{k=1}^{K} \bigg).
\end{align*}

\subsection{Proposed Aggregation Method}
\label{sub:proposed}
We start with the assumption that we have access to $K$ hallucination scores and aim to construct an improved hallucination detector using these scores. The primary challenge in aggregating these scores arises from the fact that they are generated in an unconstrained setting, meaning that each score may be measured on a different scale. Consequently, the initial step is to devise a method for standardizing these scores to enable their aggregation. The standardization weights we propose, $w_k$, are specific to each detection score. Using the min-max normalization, they are designed based on the whole training dataset $\mathcal{D}_n=\{x_1, \ldots, x_n \}$. Formally:
\begin{equation*}
    w_k = \frac{ s_k(x') - \underset{z \in \mathcal{D}_n}{\mathrm{min}} ~s_k(z)}{\underset{z \in \mathcal{D}_n}{\mathrm{max}} ~s_k(z) - \underset{z \in \mathcal{D}_n}{\mathrm{min}} ~s_k(z)}.
\end{equation*}
Given these weights, we  build a hallucination detector based on a weighted averaged of the score $s_k$ relying upon the previous ‘‘normalization weights'':
\begin{equation}
    \operatorname{Agg} (x') = \sum_{k=1}^{K} w_k s_k(x').
\label{eq:stare}
\end{equation}
We denote this method as \textsf{STARE}.

\begin{table*}[t]
\centering
\footnotesize
\renewcommand\arraystretch{1}
\setlength{\tabcolsep}{4.5pt}
\begin{subtable}[t]{0.45\linewidth}
\begin{tabular}{>{\arraybackslash}m{2.45cm} r r}
\toprule
\textsc{Detector} & \multicolumn{1}{c}{AUROC $\uparrow$} & \multicolumn{1}{c}{FPR $\downarrow$}\\
\midrule
\multicolumn{3}{c}{{\textit{Individual Detectors}}} \\\midrule
\multicolumn{3}{l}{{\textit{External}}} \\
\textsf{COMET-QE}       & 70.15 \phantom{0000} & {57.24} \phantom{0000}
\\\textsf{CometKiwi} & 86.96 \phantom{0000} & 35.15 \phantom{0000}
\\\textsf{LaBSE} & \underline{91.72}\,\,\, \faMedal & \underline{26.86}\,\,\, \faMedal
\\\cdashlinelr{1-3}\noalign{\vskip 0.5ex} 

\multicolumn{3}{l}{{\textit{Model-based}}} \\\textsf{Seq-Logprob} & 83.40 \phantom{0000} & 58.99 \phantom{0000}
\\\textsf{ALTI+} & 84.24 \phantom{0000} & 66.19 \phantom{0000} 
\\\textsf{Wass-Combo} & \underline{87.02} \phantom{0000} & \underline{48.38} \phantom{0000}
\\\midrule\midrule
\multicolumn{3}{c}{{\textit{Aggregated Detectors}}} \\\midrule
\multicolumn{3}{l}{{\textit{External Only (gap to best single External)}}} \\\textsf{Isolation Forest} & 92.61 \uag{0.89} & 19.08 \da{7.78} 
\\\textsf{Max-Norm} & 92.43 \uag{0.71} & 22.09 \da{4.77}
\\\textsf{STARE} & 93.32 \uag{1.60} & 20.67 \da{6.19}
\\\cdashlinelr{1-3}\noalign{\vskip 0.5ex} 
\multicolumn{3}{l}{{\textit{Model-based Only (gap to best single Model-based)}}} \\\textsf{Isolation Forest} & 88.19 \uag{1.17} & 36.63 \da{11.8}
\\\textsf{Max-Norm} & 83.81 \dab{3.21} & 62.94 \ua{14.6} 
\\\textsf{STARE} & 89.07 \uag{2.05} & 42.50 \da{5.88}
\\\cdashlinelr{1-3}\noalign{\vskip 0.5ex} 
\multicolumn{3}{l}{{\textit{All (gap to best overall)}}} \\\textsf{Isolation Forest} & 92.84 \uag{1.12} & 23.90 \da{2.96} 
\\\textsf{Max-Norm} & 91.60 \dab{0.12} & 26.38 \da{0.48}
\\\textsf{STARE} & \textbf{94.12} \uag{2.40} & \textbf{17.06} \da{9.80} 
\\\bottomrule
\end{tabular}
\caption{Results on \textsc{LfaN-Hall}.}
\end{subtable}
\hfill
\begin{subtable}[t]{0.45\linewidth}
\centering
\footnotesize
\begin{tabular}{>{\arraybackslash}m{2.45cm} r r}
\toprule
\textsc{Detector} & \multicolumn{1}{c}{AUROC $\uparrow$} & \multicolumn{1}{c}{FPR $\downarrow$}\\
\midrule
\multicolumn{3}{c}{{\textit{Individual Detectors}}} \\\midrule
\multicolumn{3}{l}{{\textit{External}}} \\
\textsf{COMET-QE}       & 82.22 \phantom{0000} & 47.40 \phantom{0000}
\\\textsf{LASER} & 81.11 \phantom{0000} & 47.04 \phantom{0000}
\\\textsf{XNLI} & 82.44 \phantom{0000} & \underline{33.20} \phantom{0000}
\\\textsf{LaBSE} & \underline{88.77}\,\,\, \faMedal & 34.96\,\,\, \faMedal
\\\cdashlinelr{1-3}\noalign{\vskip 0.5ex} 
\multicolumn{3}{l}{{\textit{Model-based}}} \\
\textsf{Seq-Logprob} & \underline{86.72} \phantom{0000} & \underline{28.86} \phantom{0000}
\\\textsf{ALTI+} & 82.26 \phantom{0000} & 58.40 \phantom{0000} 
\\\textsf{Wass-Combo} & 64.82 \phantom{0000} & 84.62 \phantom{0000}
\\\midrule\midrule
\multicolumn{3}{c}{{\textit{Aggregation Detectors}}} \\\midrule
\multicolumn{3}{l}{{\textit{External Only (gap to best single External)}}}
\\\textsf{Isolation Forest} & 71.35 \dab{17.4} & 57.75 \ua{22.8} 
\\\textsf{Max-Norm} & 88.57 \uag{0.48} & 32.59 \da{2.86}
\\\textsf{STARE} & 89.76 \uag{0.99} & 32.74 \da{2.22}
\\\cdashlinelr{1-3}\noalign{\vskip 0.5ex} 
\multicolumn{3}{l}{{\textit{Model-based Only (gap to best single Model-based)}}} 
\\\textsf{Isolation Forest} & 75.35 \dab{11.4} & 69.71 \ua{40.9}
\\\textsf{Max-Norm} & 67.70 \dab{17.3} & 83.83 \ua{53.1} 
\\\textsf{STARE} & 89.92 \uag{3.20} & 30.37 \ua{1.51}
\\\cdashlinelr{1-3}\noalign{\vskip 0.5ex} 
\multicolumn{3}{l}{{\textit{All (gap to best overall)}}} 
\\\textsf{Isolation Forest} & 76.25 \dab{12.5} & 56.28 \ua{21.3} 
\\\textsf{Max-Norm} & 80.67 \dab{7.01} & 41.52 \ua{1.91}
\\\textsf{STARE} & \textbf{91.18} \uag{2.41} & \textbf{28.85} \da{6.11} 
\\\bottomrule
\end{tabular}
\centering
\caption{Results on \textsc{HalOmi}.}
\end{subtable} \hspace{10pt}
\caption{Performance, according to AUROC and FPR, of all single detectors available and aggregation methods via combination of external detectors, model-based detectors, or both simultaneously. We represent with \faMedal the best overall single detector and underline the best detectors for each class, according to our primary metric AUROC.}
\label{tab:auroc_fpr_type}
\end{table*}



\section{Experimental Setup}
\subsection{Datasets}
In our experiments, we utilize the human-annotated datasets released in \citet{guerreiro2022looking} and \citet{dale2023halomi}. Both datasets include detection scores~---~both for internal and external detectors ---~for each individual translation:
\paragraph{\textsc{LfaN-Hall}.} A dataset of 3415 translations for WMT18 German$\rightarrow$English news translation data~\citep{bojar-etal-2018-findings} with annotations on critical errors and hallucinations~\citep{guerreiro2022looking}. This dataset contains a mixture of \textit{oscillatory} hallucinations and \textit{fluent but detached} hallucinations. We provide examples of such translations in Appendix~\ref{app:hall_datasets}. For each translation, there are six different detector scores: three are from external models (scores from \textsc{COMET-QE} and \textsf{CometKiwi}, two quality estimation models, and sentence similarity from \textsf{LaBSE}, a cross-lingual embedding model), and three are from internal methods (length-normalized sequence log-probability, \textsf{Seq-Logprob}; contribution of the source sentence for the generated translation according to \textsf{ALTI+}~\citep{ferrando-etal-2022-towards}, and \textsc{Wass-Combo}, an Optimal Transport inspired method that relies on the aggregation of attention maps).

\paragraph{\textsc{HalOmi}.} A dataset with human-annotated hallucination in various translation directions. We test translations into and out of English, pairing English with five other languages --- Arabic, German, Russian, Spanish, and Chinese, consisting of over 3000 sentences across the ten different language pairs. Importantly, this dataset has two important properties that differ from \textsc{LfaN-Hall}: (i)~it has a much bigger proportion of fluent but detached hallucinations~(oscillatory hallucinations were not considered as a separate category), and (ii)~nearly 35$\%$ of the translations are deemed hallucinations, as opposed to about 8$\%$ for \textsc{LfaN-Hall}.\footnote{Given the rarity of hallucinations in practical translation scenarios~\citep{10.1162/tacl_a_00615}, \textsc{LfaN-Hall} offers a more realistic simulation of detection performance.} For each translation, there are seven different detection scores: the same internal detection scores as \textsc{LfaN-Hall}, and four different detector scores: \textsf{COMET-QE}, \textsf{LASER}, \textsf{XNLI} and \textsf{LaBSE}.

We provide more details on both datasets in Appendix~\ref{app:hall_datasets}.

\paragraph{Aggregation Baselines.} The closest related work is \citet{darrin2023unsupervised} on out-of-distribution detection methods, using an \textsf{Isolation Forest} (IF; \citealp{liu2008isolation}) for per-class anomaly scores. We adapt their method, employing a single \textsf{Isolation Forest}, and designate it as our baseline. Alternatively, we also consider a different way to use the individual scores and normalization weights in Equation~\ref{eq:stare}: instead of performing a sum over the weighted scores, we take the maximum score. We denote this baseline as \textsf{Max-Norm}.

\paragraph{Evaluation method.} Following \citet{guerreiro2022optimal}, we report Area Under the Receiver Operating Characteristic curve (AUROC) as our primary metric, and False Positive Rate at 90\% True Positive Rate (FPR@90TPR) as a secondary metric.

\paragraph{Implementation details.} For \textsc{LfaN-Hall}, we normalize the metrics by leveraging the held-out set released with the dataset consisting of 100,000 non-annotated in-domain scores. In the case of \textsc{HalOmi}, however, no held-out set was released. As such, we rely on sampling random splits that consist of 10\% of the dataset for calibration. We repeat the process 10 different times. We report average scores over those different runs. We also report the performance variance in the Appendix. 


\subsection{Performances Analysis}
Results on hallucination detection performance on \textsc{LfaN-Hall} and HaloMNI are reported in \autoref{tab:auroc_fpr_type}.

\paragraph{Global Analysis.} \textsf{STARE} aggregation method consistently outperforms (i) single detectors' performance, and (ii) other aggregation baselines. Moreover, we find that the combination of all detectors --- both model-based and external-based detectors --- yields the best overall results, improving over the \textsf{STARE} method based on either internal or external models only. Importantly, these trends, contrary to other alternative aggregation strategies, hold across both datasets.

\paragraph{Aggregation of External Detectors.} \textsf{STARE} demonstrates robust performance when aggregating external detectors on both \textsc{LfaN-Hall} and \textsc{HalOmi}: improvements in AUROC (over a point) and in FPR (between two to six points). Interestingly, we also observe that the best overall performance obtained exclusively with external models lags behind that of the overall aggregation. This suggests that internal models features --- directly obtained via the generation process --- contribute with complementary information to that captured by external models.

\paragraph{Aggregation of Internal Detectors.} Aggregation of internal detectors, can achieve higher AUROC scores than the best single external detector on \textsc{HalOmi}. This results highlights how model-based features --- such as attention and sequence log-probability --- that are readily and efficiently obtained as a by-product of the generation can, when aggregated effectively, outperform more computationally expensive external solutions.

\subsection{Ablation Studies}\label{sec:ablation}
In this section, our focus is two-fold: (i)~exploring optimal selections of detectors, and (ii)~understanding the relevance of the reference set's size.

\paragraph{Optimal Choice of detectors.} We report the performance of the optimal combination of $N$-detectors on both datasets in \autoref{tab:optimal}.\footnote{We report the optimal combinations in Appendix~\ref{app:optimal_combination_of_detectors}.} We note that including all detectors yields comparable performance to the best mix of detectors. Interestingly, aggregation always brings improvement, even when only combining two detectors. As expected, the best mixture of detectors leverages information from different signals: contribution of source contribution, low-quality translations, and dissimilarity between source and translation.

\begin{figure}
    \centering
     \includegraphics[width=0.22\textwidth]{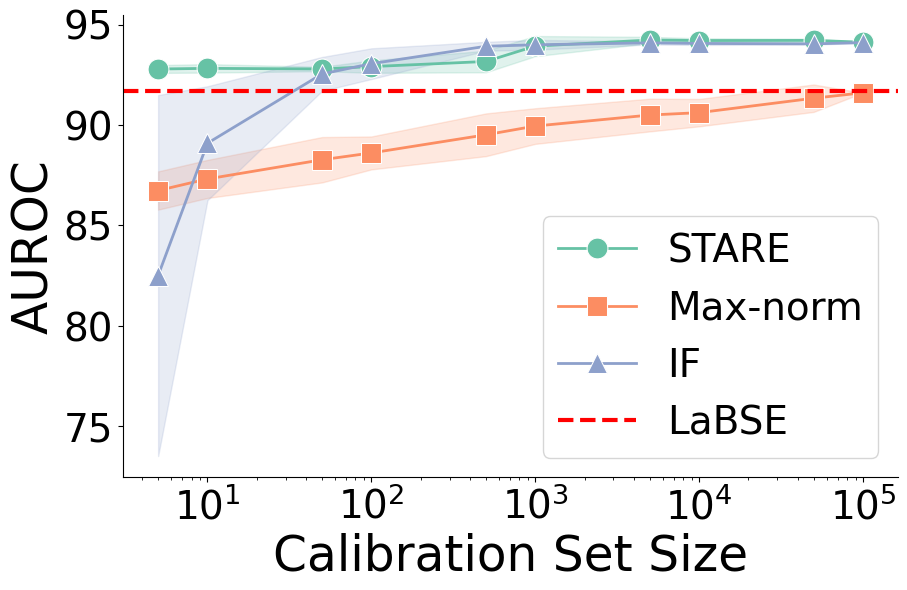}
    \includegraphics[width=0.22\textwidth]{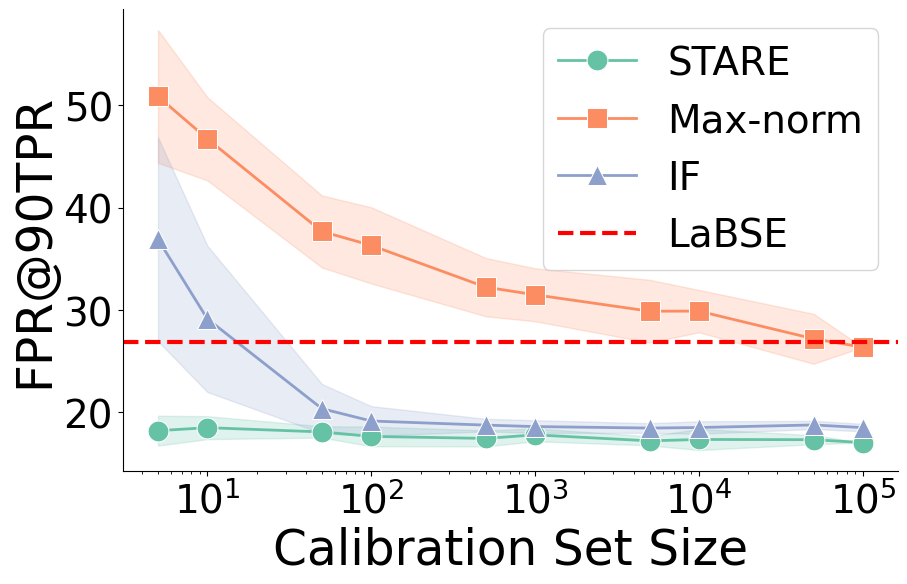}
    \caption{Impact of reference set size on \textsc{LfaN-Hall}.}
    \label{fig:ablation_reference}
\end{figure}

\renewcommand{\arraystretch}{0.85}
\begin{table}[h]
    \centering
    \resizebox{\linewidth}{!}{
    \begin{tabular}{ccccc}\toprule
     & \multicolumn{2}{c}{\textsc{LfaN-Hall}} & \multicolumn{2}{c}{\textsc{HalOmi}} \\\cdashlinelr{1-5}
        $N$ & AUROC & FPR@90 & AUROC & FPR@90 \\\midrule
        \textsf{LaBSE} & 91.72 & 26.86 & 88.77 & 34.96\\\cdashlinelr{1-5}
        2 & 93.32 & 20.67 &90.40&27.52\\
        3 & 94.11 & 17.27 &90.61&27.24\\
        4 & 94.45 & 13.69 &91.09&26.91\\
        5 & 94.12 & 17.06 &91.25&28.48\\
        6 & --- & --- &91.40&27.93\\\cdashlinelr{1-5}
        \textsf{STARE} & 94.12 & 17.06 & 91.18 & 28.85\\\bottomrule
    \end{tabular}
    }
    \caption{Ablation Study on the Optimal Choice of Detectors when using \textsf{STARE}.}
    \label{tab:optimal}
\end{table}
\paragraph{Impact of the size of the references set.} The calibration of scores relies on a reference set. Here, we examine the impact of the calibration set size on performance, by ablating on the held-out set \textsc{LfaN-Hall}, which comprises of 100k sentences. \autoref{fig:ablation_reference} shows that the \textsc{Isolation Forest} requires a larger calibration set to achieve similar performance. This phenomenon might explain the drop in performance observed on \textsc{HalOmi}~(Table~\ref{tab:auroc_fpr_type}). Interestingly, the performance improvement for \textsf{STARE}, particularly in FPR, plateaus when the reference set exceeds 1,000 samples, which suggests that \textsf{STARE} can adapt to different domains with a rather small reference set.

\section{Conclusion \& Future Perspectives}
We propose a simple aggregation method to combine hallucination detectors to exploit complementary benefits from each individual detector. We show that our method can bring consistent improvements over previous detection approaches in two human-annotated datasets across different language pairs. We are also releasing our code and detection scores to support future research on this topic.

\clearpage
\section{Limitations}
Our methods are evaluated in a limited setup due to the limited availability of translation datasets with annotation of hallucinations. Moreover, in this study, we have not yet studied \textit{compute-optimal} aggregation of detectors --- we assume that we already have access to multiple different detection scores.
\section{Acknowledgements}
Training compute is obtained on the Jean Zay supercomputer operated by GENCI IDRIS through compute grant 2023-AD011014668R1, AD010614770 as well as on Adastra through project c1615122, cad15031, cad14770 .

\bibliography{anthology,custom}
\bibliographystyle{acl_natbib}
\newpage
\clearpage
\appendix
\section{Model and Data Details}
\label{app:hall_datasets}
\subsection{\textsc{LfaN-Hall} dataset}
\label{app:lfan_hall_dataset}
\paragraph{NMT Model.} The model used in \citet{guerreiro2022looking} is a Transformer base model~\cite{transformer_vaswani}~(hidden size of 512, feedforward size of 2048, 6 encoder and 6 decoder layers, 8 attention heads). The model has approximately 77M parameters. It was trained on WMT18 \textsc{de-en} data: the authors randomly choose 2/3 of the dataset for training and use the remaining 1/3 as a held-out set for analysis. We use a section of that same held-out set in this work.

\paragraph{Dataset Stats.} The dataset consists of 3415 translations from WMT18 \textsc{de-en} data. Overall, there are 218 translations annotated as detached hallucinations (fully and strongly detached --- see more details in~\citet{guerreiro2022looking}), and 86 as oscillatory hallucinations.\footnote{Some strongly detached hallucinations have also been annotated as oscillatory hallucinations. In these cases, we follow~\citet{guerreiro2022optimal} and consider them to be oscillatory.} The other translations are either incorrect (1073) or correct (2048). We show examples of hallucinations for each category in Table~\ref{tab:hallucinationexamples}.\footnote{All data used in this paper is licensed under a MIT License.}

\begin{table}[t]
\centering
\footnotesize
\renewcommand\arraystretch{1}
\setlength{\tabcolsep}{4.5pt}
\begin{tabular}{>{\arraybackslash}m{2.45cm} r r}
\toprule
\textsc{Detector} & \multicolumn{1}{c}{AUROC $\uparrow$} & \multicolumn{1}{c}{FPR@90TPR $\downarrow$}\\
\midrule
\multicolumn{3}{c}{{\textit{Individual Detectors}}} \\\midrule
\multicolumn{3}{l}{{\textit{External}}} \\
\textsf{COMET-QE}       & 82.22 $\pm$ 0.28 & 47.40 $\pm$ 0.82 \\
\textsf{LASER}           & 81.11 $\pm$ 0.21 & 47.04 $\pm$ 0.78 \\
\textsf{XNLI}            & 82.44 $\pm$ 0.18 & 33.20 $\pm$ 0.63 \\
\textsf{LaBSE}           & 88.77 $\pm$ 0.21 & 34.96 $\pm$ 0.72 \\
\cdashlinelr{1-3}\noalign{\vskip 0.5ex} 
\multicolumn{3}{l}{{\textit{Model-based}}} \\
\textsf{Seq-Logprob}       & 86.72 $\pm$ 0.22 & 28.86 $\pm$ 0.64 \\
\textsf{ALTI+}      & 82.26 $\pm$ 0.28 & 58.40 $\pm$ 0.54 \\
\textsf{Wass-Combo}        & 64.82 $\pm$ 0.20 & 84.62 $\pm$ 0.52 \\
\midrule\midrule
\multicolumn{3}{c}{{\textit{Aggregated Detectors}}} \\\midrule
\multicolumn{3}{l}{{\textit{External Only}}} \\
\textsf{Isolation Forest}       & 71.35 $\pm$ 1.62 & 57.75 $\pm$ 4.55 \\
\textsf{Max-Norm}               & 88.57 $\pm$ 0.38 & 32.59 $\pm$ 0.60 \\
\textsf{STARE}               & 89.76 $\pm$ 0.19 & 32.74 $\pm$ 0.50 \\
\cdashlinelr{1-3}\noalign{\vskip 0.5ex} 
\multicolumn{3}{l}{{\textit{Model-based Only}}} \\
\textsf{Isolation Forest}       & 75.35 $\pm$ 2.32 & 69.71 $\pm$ 5.01 \\
\textsf{Max-Norm}               & 67.70 $\pm$ 1.31 & 83.83 $\pm$ 1.40 \\
\textsf{STARE}               & 89.92 $\pm$ 0.20 & 30.37 $\pm$ 1.84 \\
\cdashlinelr{1-3}\noalign{\vskip 0.5ex} 
\multicolumn{3}{l}{{\textit{All}}} \\
\textsf{Isolation Forest}       & 76.25 $\pm$ 2.16 & 56.28 $\pm$ 6.29 \\
\textsf{Max-Norm}               & 80.67 $\pm$ 1.37 & 41.52 $\pm$ 5.87 \\
\textsf{STARE}               & 91.18 $\pm$ 0.20 & 28.85 $\pm$ 0.89 \\
\bottomrule
\end{tabular}
\caption{Performance of individual and aggregated hallucination detectors on the \textsc{HalOmi} dataset, including average performance and standard deviations across ten different calibration sets.}
\label{app:tab_w_variance}
\end{table}

\begin{table*}[t]
\centering
\scriptsize
\renewcommand\arraystretch{1.0}
\begin{tabular}{
>{\arraybackslash}m{1.0cm} >{\arraybackslash}m{4.8cm}
>{\arraybackslash}m{4.5cm}
>{\arraybackslash}m{4cm}}
\toprule
\textbf{Category} &  \textbf{Source Sentence} & \textbf{Reference Translation} & \textbf{Hallucination}\\ \midrule
Oscillatory & Als Maß hierfür wird meist der sogenannte Pearl Index benutzt (so benannt nach einem Statistiker, der diese Berechnungsformel einführte). & As a measure of this, the so-called Pearl Index is usually used (so named after a statistician who introduced this calculation formula). & The \textcolor{HallRedText}{\hl{term "Pearl Index"}} refers to \textcolor{HallRedText}{\hl{the term "Pearl Index"}} (or \textcolor{HallRedText}{\hl{"Pearl Index"}}) used to refer to \textcolor{HallRedText}{\hl{the term "Pearl Index"}} (or \textcolor{HallRedText}{\hl{"Pearl Index"}}).\\[1.25ex] \midrule
Strongly Detached & Fraktion der Grünen / Freie Europäische Allianz & The Group of the Greens/European Free Alliance & \textcolor{HallRedText}{\hl{Independence and Democracy}} Group \textcolor{HallRedText}{\hl{(includes 10 UKIP MEPs and one independent MEP from Ireland)}} \\[1.25ex] \midrule
Fully \ \ \ \ \ \  Detached & Die Zimmer beziehen, die Fenster mit Aussicht öffnen, tief durchatmen, staunen. & Head up to the rooms, open up the windows and savour the view, breathe deeply, marvel. &  \textcolor{HallRedText}{\hl{The staff were very friendly and helpful.}} \\ 
\bottomrule
\end{tabular}
\caption{Examples of hallucination types. Hallucinated content is shown \textcolor{HallRedText}{\hl{shaded}}.}
\label{tab:hallucinationexamples}
\end{table*}

\subsection{\textsc{HalOmi} dataset}
\paragraph{NMT model.} Translations on this dataset come from 600M distilled NLLB model~\citep{nllb2022}.

\section{Variance of performance on the \textsc{HalOmi} dataset}
We report in Table~\ref{app:tab_w_variance} the average performance as well as the standard deviation across the different ten runs on different calibration sets. Despite variance between different runs, the STARE aggregation method consistently outperforms individual detectors and other aggregation techniques.

\section{Optimal Combination of Detectors via \textsf{STARE}}
\label{app:optimal_combination_of_detectors}
\paragraph{\textsc{LfaN-Hall}.} The optimal set of detectors for various values of \( N \) is:
\begin{itemize}
\item for \( N=1 \): \textsf{LaBSE}
\item for \( N=2 \): \textsf{CometKiwi}, \textsf{LaBSE}
\item for \( N=3 \): \textsf{Wass\_Combo}, \textsf{CometKiwi}, \textsf{LaBSE}
\item for \( N=4 \): \textsf{ALTI+}, \textsf{Wass\_Combo}, \textsf{CometKiwi}, \textsf{LaBSE}
\item for \( N=5 \): \textsf{ALTI+}, \textsf{SeqLogprob}, \textsf{Wass\_Combo}, \textsf{CometKiwi}, \textsf{LaBSE}
\end{itemize}

\paragraph{\textsc{HalOmi}.} The optimal set of detectors for various values of \( N \) is:
\begin{itemize}
    \item for \( N=2 \): \textsf{LaBSE}, \textsf{SeqLogprob}
    \item for \( N=3 \): \textsf{LaBSE}, \textsf{SeqLogprob}, \textsf{Wass-Combo}
    \item for \( N=4 \): \textsf{LaBSE}, \textsf{SeqLogprob}, \textsf{XNLI}, \textsf{COMET-QE}
    \item for \( N=5 \): \textsf{LaBSE}, \textsf{SeqLogprob}, \textsf{XNLI}, \textsf{COMET-QE}, \textsf{ALTI+}
    \item for \( N=6 \): \textsf{LaBSE}, \textsf{Log Loss}, \textsf{XNLI}, \textsf{COMET-QE}, \textsf{ALTI+}, \textsf{Wass-Combo}
    \item for \( N=7 \): \textsf{LaBSE}, \textsf{SeqLogprob}, \textsf{XNLI}, \textsf{COMET-QE}, \textsf{ALTI+}, \textsf{Laser}, \textsf{Wass-Combo}
\end{itemize}

\section{Future Work \& Perspectives}
In the future, we would like to explore more anomaly detection methods to improve the aggregation quality. Specifically, we would like to test Information Projections \cite{picot2023simple,darrin2023rainproof} and data depths \cite{picot2023adversarial,picot2023halfspace,colombo2022beyond,staerman2021pseudo,colombo2023toward}.

\end{document}